\documentclass[letterpaper, 10 pt, conference]{utility/ieeeconf} 

\IEEEoverridecommandlockouts  

\usepackage{amssymb}
\usepackage{amsmath}
\usepackage{graphicx} 
\usepackage{algorithm} 
\usepackage{algpseudocode} 
\usepackage{url} 
\usepackage{hyperref} 
\usepackage{cite}  

\title{\LARGE \bf
Image-Based Deep Reinforcement Learning with Intrinsically Motivated Stimuli: On the Execution of Complex Robotic Tasks
}

\author{David Valencia, Henry Williams, Yuning Xing, Trevor Gee, Minas Liarokapis, Bruce A. MacDonald%
\thanks{D. Valencia, H. Williams, Y. Xing,  T. Gee, and B. MacDonald are with the Centre for Automation and Robotic Engineering Science, The University of Auckland, New Zealand. Emails: \{dval035,yxin683\}@aucklanduni.ac.nz, \{t.gee,b.macdonald,henry.williams\}@auckland.ac.nz}
\thanks{M. Liarokapis is with the New Dexterity Research Group, The University of Auckland, New Zealand. Email: minas.liarokapis@auckland.ac.nz}%
}

\date{September 2023}

\begin{document}
\maketitle

\begin{abstract}
    Reinforcement Learning (RL) has been widely used to solve tasks where the environment consistently provides a dense reward value. However, in real-world scenarios, rewards can often be poorly defined or sparse. Auxiliary signals are indispensable for discovering efficient exploration strategies and aiding the learning process. In this work, inspired by intrinsic motivation theory, we postulate that the intrinsic stimuli of novelty and surprise can assist in improving exploration in complex, sparsely rewarded environments. We introduce a novel sample-efficient method able to learn directly from pixels, an image-based extension of TD3  with an autoencoder called \textit{NaSA-TD3}. The experiments demonstrate that NaSA-TD3 is easy to train and an efficient method for tackling complex continuous-control robotic tasks, both in simulated environments and real-world settings. NaSA-TD3 outperforms existing state-of-the-art RL image-based methods in terms of final performance without requiring pre-trained models or human demonstrations.
\end{abstract}

\section{Introduction} 

    Motivation could be seen as an internal force or energy to encourage someone to execute or avoid a task.
    It is related to several internal or external factors, such as personality, own beliefs, culture, the surrounding environment, or earlier experiences.
    Motivation permits agents, organisms such as an animal or artificial organisms such as a robot, to play, explore, and discover new behaviours without external reward signals \cite{barto2013intrinsic, ryan2000self, aubret2023information}. 
    Psychologists distinguish two groups of motivation, intrinsic and extrinsic \cite{ryan2000self,siddique2017review}. When you do something for its satisfaction, i.e. without any immediate obvious intention, it is considered intrinsic motivation \cite{baldassarre2019intrinsic, aubret2023information}. In contrast,  extrinsic motivation is when you do something to obtain positive feedback associated with that behaviour and the environment.
    Inside the intrinsic motivation group, novelty and surprise play an essential role in the agent's learning process. They stimulate interest and exploration of new stimuli \cite{baldassarre2013intrinsically, barto2013novelty}. As mentioned in \cite{ryan2000self}, only novel, unexpected and challenging circumstances make an organism intrinsically motivated. Specialized literature  \cite{schomaker2019unexplored, jonassaint2007facets, schomaker2015short, swan1996curiosity} claims that exploration of new environments enriches plasticity in the brain to stimulate learning. This phenomenon is easy to recognize in young human infants, who show significant preferences for situations or stimuli never experienced before, even when these stimuli do not provide an associated direct reward \cite{jaegle2019visual, reynolds2015infant}. 
    
    The underlying idea is simple: a novel situation or a surprise emotion encourages the agent to explore and learn better solutions faster than an agent using only external reward signals \cite{baldassarre2013intrinsically,barto2013novelty,krebs2009novelty,duzel2010novelty,guitart2010contextual}. This is particularly important in the real world, especially in robotics, where most basic tasks do not have an immediate external reward or are absent. Furthermore, not only are intrinsic and extrinsic reward signals required to learn to solve tasks, but rich and abundant information is also needed, information that can describe and provide key features of the environment and the task itself.  This is particularly important in RL applications, where the state space representation plays a critical role during the learning process. 
    
    Learning from images has proven to be a feasible choice in RL \cite{yarats2021improving, lee2020stochastic, hafner2019learning, stadie2015incentivizing}. Images encapsulate valuable information that may not always be readily available in other representations, such as proprioceptive states of the dynamics. Moreover, learning directly from pixels offers scalability for real-world applications. Utilizing cameras to gather environmental information proves to be more accurate, cost-effective, and straightforward compared to other sensor types, especially in complex setups featuring high dynamics, such as robots with multiple degrees of freedom (DoF).
    
    Nonetheless, learning directly from high-dimensional images presents a considerable challenge. Most of the state-of-the-art RL algorithms struggle when the input state is pure pixels \cite{yarats2021improving}, as well as the computational cost increase \cite{stadie2015incentivizing}. Thus, having an accurate but simplified compact representation of the image that the agent can use to learn a control policy might help reduce the complexity and cost. However, creating a compact and valuable representation of an image with all relevant information about the environment and the task from which to learn is a challenge \cite{lee2020stochastic}. 
    
    Mapping images into a useful encoding usually demands a large training dataset, which is not a particular feature in RL. Ideally, mapping the images into a compressed representation and learning a control policy should co-occur and depend on each other. This interdependence resembles a chicken-and-egg dilemma: an optimal policy necessitates an effective representation, while an effective representation relies on meaningful gradient information derived from the policy. Therefore, achieving synergy between representation learning and policy optimization is paramount.
    
    Thus, in this study, aiming to address the challenges of sparse reward environments and image-based learning representations, we introduce a novel RL algorithm called Novelty and Surprise Autoencoder-TD3, \textbf{NaSA-TD3}. We extend and adapt the TD3 framework proposed by \cite{fujimoto2018addressing} from a conventional deterministic actor-critic method to a comprehensive image-based actor-critic approach. Drawing inspiration from intrinsic motivation theory, we translate human motivational behaviours into a computational framework by representing novel and surprising stimuli. We leverage these stimuli to uncover more effective exploration strategies to enhance the learning process. Additionally, to support and facilitate the endeavour of learning directly from images, we employ an autoencoder network in conjunction with an actor-critic methodology. This combination enables learning control policies tailored for executing complex robotic tasks, such as dexterous manipulations. The paper's contributions can be summarized as follows:
    
    \begin{enumerate}
        \item A computational representation of human intrinsic stimuli novelty and surprise.  
        \item Evaluation of the influence of intrinsic stimuli novelty and surprise, introduced as a reward bonus, on RL for complex robotics tasks.
        \item  An image-based extension of TD3 \cite{fujimoto2018addressing} with an autoencoder for learning control policies for continuous action problems.
    \end{enumerate}
    
\section{Definitions} 
    Novelty and surprise are closely related to each other. Their definitions are often incorrectly interchanged or interpreted in the same way; however, there are precise attributes that make them distinguishable \cite{barto2013novelty, xu2021novelty}. 
    
    It is important to mention that there are several other detailed concepts relevant to motivation and intrinsic stimulus from a psychological perspective, along with extensive neuroscience literature; however, they are beyond this article's scope, and we do not address them because it would put us in a category too far from our primary goal. Nevertheless, for more in-depth definitions, we encourage the reader to review \cite{oudeyer2008can} and \cite{oudeyer2007intrinsic}. This work uses the following definitions.
    
\subsection{ What is Novelty?}

    The Oxford  English Dictionary defines novelty as \emph{``the quality of being new, different, and interesting"} \cite{oxforddic}. Psychologists, psychiatrists, and scientists differ in describing and measuring novelty depending if the stimuli are physical, visual, or behavioural \cite{siddique2017review}. In all cases, however, the common word is \emph{\textbf{different}} \cite{siddique2017review}. We define novelty as something that we have not seen before; in other words, the process of recognizing new stimuli that are distinguishable from anything known before. 
    Detecting novelty involves digging into the contents of the agent's memory to associate or recognize if an experience has previously been seen or attended. Consequently, whether an observation is novel depends on whether a representation of that observation can be found, partial or identical, in the memory \cite{ aubret2023information, barto2013novelty}.
    For instance, in dexterous manipulation using a robotic gripper, a novel stimulus might manifest as a joint configuration or object state that the agent has not previously encountered.
    
\subsection{What is Surprise?}

    A surprise could be defined as a \textit{\textbf{discrepancy}}, presented as a feeling or emotion excited, between an expected observation and an actual observation \cite{barto2013novelty, xu2021novelty}. As mentioned in \cite{becker2021exploration}, \emph{``Surprise requires an internal world model that formulates an expectation about the future"}. Measuring surprise involves a mechanism of prediction and comparison between expectation and actual observation.  Besides, here we do not examine the current observation with past observations; i.e. surprise does not directly correlate with the memory, even though a predicted world model is built using previous experiences \cite{barto2013novelty}. 
    In dexterous manipulation with a robot gripper, a surprise stimulus can be conceptualized as the disparity between the expected position of the gripper and the target and the actual positions after taking an action. With surprise, time is important since prediction involves a specific time based on the current observation \cite{barto2013novelty}, while novelty can not be associated with predictions since we can not predict situations or events never been experienced \cite{schomaker2019unexplored, schomaker2015short}. 
    
\section{Related work}

    The idea of using intrinsic motivation has been used in RL to encourage the agent's exploration. 
    In low-domain environments, a common method for computing exploration bonus signals is having a tabular record of how often a state has been visited, i.e. a count-based strategy. Previous proposals using count-based approaches, such as \cite{schulman2015trust, kolter2009near, oh2015action, tang2017exploration, bellemare2016unifying,brafman2002r,kearns2002near, ostrovski2017count}, have shown satisfactory results and demonstrated tackling the exploration problem. However, maintaining state-action visitation count is impractical, inefficient and expensive in complex tasks with continuous state spaces,  making count-based approaches unscalable \cite{jaegle2019visual}. For example, it would be impractical in a robotics application to try to save all possible states when we deal with an infinite state space. 
    
    To avoid a tabular record of every visited state, \cite{fu2017ex2} proposes EX2, a pseudo-count exploration method based on a discriminative exemplar model where the authors use a K-exemplar conditioned network. This approach, however, acts as a binary classifier that may tend to overfit due to the nature of the classifier itself. In \cite{hafez2019deep}, the authors present a method using self-organizing maps, a CNN autoencoder and a predictive ensemble model to estimate an intrinsic reward. This method was tested in a simple environment using a single actuator for a learning-to-reach task. While work by \cite{bougie2021fast} presents a concept that uses an agent's ability to reconstruct observations by giving their ``contexts" (a downsample or noise version of the input image) to generate an intrinsic bonus that is decomposed into fast and slow rewards signals. Due to the need to balance the importance of the intrinsic signals, this method can only be applied to on-policy RL methods. No experiments in real-world or continuous action spaces are presented. Learning without extrinsic rewards has also been examined in \cite{lehman2011abandoning}, where the novelty of a particular state is calculated as the distance of the current behaviour features to the k-nearest neighbours in a memory of behaviour features. However, the agent can end in ``procrastination" on the main task instead. 
    
    On the other hand, \cite{pathak2017curiosity} proposes a curiosity-based method to help the agent explore the environment more by generating an intrinsic signal based on an inverse dynamics model. In this proposal, the authors transform the observation input into a feature space with the relevant information and only consider the predictions that can affect the agent and ignore the rest. In \cite{bougie2020skill}, a goal-based curiosity method is presented, where the authors decompose a task into several easier sub-tasks (called goals) given an observation. The agent samples K goals and predicts the probability that each goal can be achieved from the current observation using a predictor network. The number of samples needed to train the policy is too high to scale this method to real-world setups.

    Most closely related to our method is the work presented in \cite{aljalbout2022seeking}, which introduces a curiosity-based approach coupled with a comprehensive analysis of state representation learning (SRL) to enhance the sample efficiency of image-based RL. In this study, two methods are proposed: one utilizing a regularized autoencoder (RAE), and the other employing a contrastive learning approach. However, this work differs from ours in that it does not integrate intrinsic signals as part of the reward; instead, a separate curious policy based on the SRL error is trained to foster interest in problematic states. This approach is applicable only to off-policy methods and has yet to be validated in real-world environments.
    
    Despite the promising results demonstrated in these studies, intrinsic motivation remains underutilized as a potential solution in RL. While related works have illustrated that employing intrinsic stimuli as bonus rewards or additional policy promotes agent exploration, only a subset of these approaches are viable for real-world deployment due to limitations in sample efficiency or computational cost. Moreover, many related works have been evaluated in simplistic real-world environments or simulations, often relying on intricate calculations, predefined initial conditions, or pre-trained models \cite{lee2020stochastic}. Addressing the majority of these challenges, we present a scalable and robust method capable of tackling control tasks in continuous action spaces. Our algorithm leverages human-inspired stimuli of novelty and surprise to enhance RL learning directly from images. In comparison to traditional heuristic approaches, our method has the ability to solve complex tasks across both simulated and real-world environments.
\section{NaSA-TD3} 
    The goal of an agent in RL is to find the best action or decision for each state to maximize future rewards $R$; nevertheless, in some complex tasks, the reward $R$  may not be present or is sparse, complicating the search for an optimal policy. However, by employing intrinsic motivation stimuli, i.e. augmenting the reward function to deliver a bonus for visiting interesting, novel or surprising states, we can encourage exploration and improve the policy in an environment where the reward is sparse. Thus, intrinsic stimuli encourage the agent to explore more than an agent using extrinsic reward signals only. 

       \begin{figure}[ht]
    \centering
        \includegraphics[width=\linewidth]{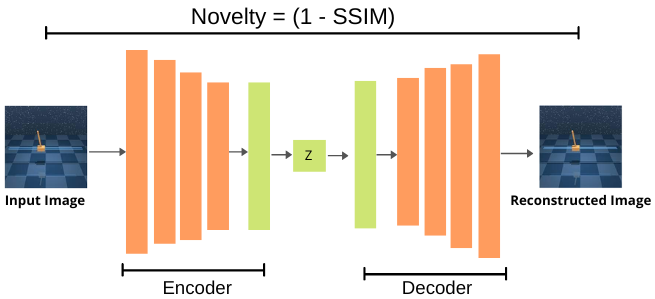}
        \caption{ Novelty detection diagram. At each time step, an observation is passed to the encoder. The decoder receives the $z$ latent presentation and reconstructs the original observation. SSIM is calculated between the reconstruction and the original observation.}
        \label{fig:Novelty}
    \end{figure}
    
    The total $R$ signal given a particular state $s$ and an action $a$ can be presented as:
    \begin{align}
        R_{\text{total}}(s, a) &= R_{\text{ext}} + R_{\text{int}} \\
        R_{\text{int}}(s, a)   &= \alpha * R_{\text{novel}} + \beta * R_{\text{surprise}}
    \end{align}
    where $ R_{ext}$ is the extrinsic reward from the environment, $R_{int}$ is the intrinsic reward formed by $R_{novel}$ and $R_{surprise}$, which are functions designed to capture the novelty and surprise of a given state-action pair and $\alpha$ and $\beta$ are the weights to balance the degree each term. In our experiments, we fix $\alpha =1$ and $\beta=1$.

\subsection{Novelty and Surprise Detection}
    
    Studies \cite{jaegle2019visual, brady2008visual, standing1973learning} have demonstrated that even after viewing a thousand pictures, each for a moment and single exposure, we humans can remember with high precision details we have seen in the picture. Recognizing whether we have seen an image is at some degree equivalent to detecting visual novelty, i.e. if we cannot remember seeing something, it can be considered a novel stimulus. As presented in \cite{jaegle2019visual}, a familiarity memory system involves the medial temporal lobe, perirhinal cortex, and inputs from the visual system to learn and recognize objects without paying attention to details. Consequently, we do not need to look for identical aspects or details in each image. Inspired by this idea, we present a novelty detection method using an unsupervised representation learning with images and an Autoencoder (AE). We determine if an observation looks ``familiar" to what we have seen previously, mimicking the human brain. 

                \begin{figure}[ht]
    \centering
        \includegraphics[width=\linewidth]{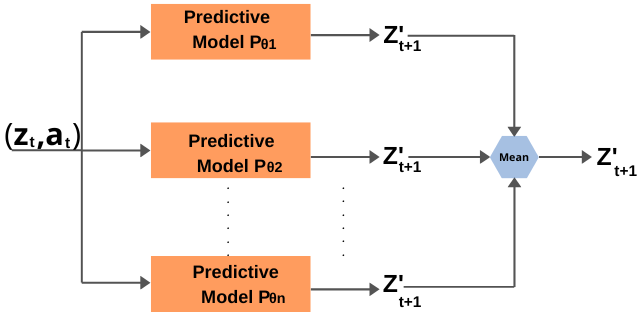}
        \caption{Ensemble of Predictive Model Architecture. Each model predicts the next $z_{t+1}$ latent presentation then the mean of the prediction is calculated.}
        \label{fig:Predictive}
    \end{figure}

    At each step, we pass the current agent's observation (an RGB image from the camera) into the AE. If the output of the AE, representing a reconstruction of the observation, deviates significantly from the original input, we infer that the observation is novel and has not been encountered previously. Although we do not explicitly compare the input image with every image stored in memory, we postulate that the AE inherently possesses a memory representation of all previously seen states within its internal layers. This implicit memory mechanism reduces computational costs and eliminates the necessity of comparing each incoming state with the entire stored memory. To quantify novelty, we propose calculating the Structural Similarity Index (SSIM) between the input observation and its reconstruction from the AE, as illustrated in Fig \ref{fig:Novelty}. Essentially, we measure the disparity between the reconstructed and original images, utilizing this difference as the intrinsic reward bonus. Our choice of SSIM stems from its ability to identify similarities within pixels with better consistency compared to other metrics, such as Mean Square Error (MSE) or Peak Signal to Noise Ratio (PSNR), which are commonly used. This selection sets our approach apart from previous works, such as \cite{sara2019image, bougie2021fast, aljalbout2022seeking}. Thus, at each time step for each observation $s_t$,  the intrinsic reward signal for novelty is calculated by:    %
     \begin{equation}
        R_{\text{novel}} (s_t) = 1 - SSIM (s_t, Dec(Enc(s_t))
    \end{equation}
    %
    Likewise, we aim to encourage agent exploration by providing a bonus signal for surprise events, i.e., to promote curiosity in unexpected situations. We measure surprise as the discrepancy between what is expected by the agent and what is actually obtained. In other words, the difference between the predicted observation and the true observation. As mentioned earlier, surprise requires an internal world model; however, predicting and generating high-dimensional observations, e.g. images, can be difficult and expensive.  Instead, we use a model to predict the next encoded image, i.e. the latent representation. This simplifies the computational cost compared to the full image reconstruction, presenting another distinctive aspect of our approach compared to previous proposals.
    
    From this, $z_{t+1}$ denotes the encoding of the image state $s_{t+1}$, and $z'_{t+1}$ the predicted representation of the next encoding state given the previous encoding image state $z_t$ and action $a_\pi$ using a predictive model $P_{\rho}$. The intrinsic reward signal for surprise is calculated by:  
    \begin{align}
        R_{\text{surprise}}(s_t, a_t) &= MSE(z_{t+1}, z'_{t+1})\\ 
        z'_{t+1}               &= P_{\rho}(z_t, a_t)
    \end{align}
    Additionally, inspired by our previous work \cite{valencia2023comparison}, we use an ensemble of predictive models. This simple but effective combination helps improve the predictions and keeps stability \cite{janner2019trust, chua2018deep} and makes the model less sensitive to noise, especially in the real world. Fig \ref{fig:Predictive} illustrates the configuration of the predictive model.

    \begin{figure}[ht]
    \centering
        \includegraphics[width=\linewidth]{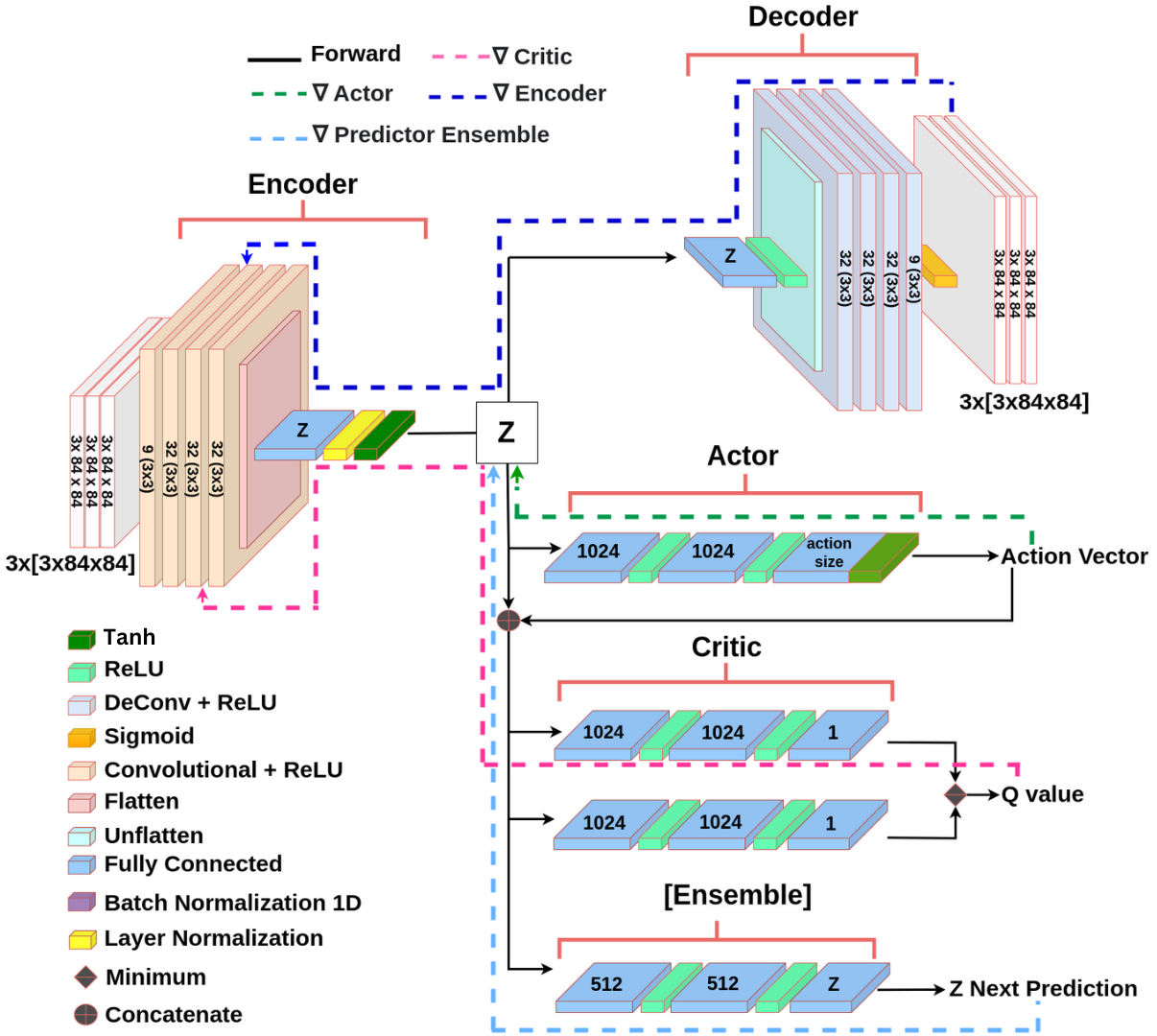}
        \caption{The proposose NaSA-TD3 method architecture.
        The encoder network consists of four convolutional layers with $32$ filters with a kernel size of $3\times3$ and $ReLU$ as an activation function.  The output of the convolutional layer is flattened and routed to a fully connected layer and a normalization layer with a $Tanh$ activation function. The Decoder network is a deconvolutional mirror of the Encoder with $Sigmoid$ as the final activation function. The TD3 network consists of an actor network and two critic networks.  All three networks have two hidden fully connected layers with $1024$ nodes each with $ReLU$. The actor has $Tanh$ as an activation function for the output layer.  The predictive ensemble model has two hidden layers with $512$ nodes and $ReLU$, and the output layer has the size of the latent $z$ vector. 
    }
        \label{fig:Architecure}
    \end{figure}
    
    It is important to underscore the distinctiveness in our approach regarding intrinsic signals and their respective computation methodologies. While previous studies, such as \cite{aljalbout2022seeking}, commonly adopt a unified intrinsic signal strategy, our research clearly emphasises delineating between novelty and surprise, employing separate methodologies for their assessment.
    
    
\subsection{Image-based policy learning}

    Inspired by \cite{fujimoto2018addressing, yarats2021improving, finn2015learning}, we introduce our Auto-Encoder Actor-Critic Off-Policy method. Our approach builds upon TD3 \cite{fujimoto2018addressing} and adheres to its training dynamics. The selection of TD3 is motivated by its simplicity and stability during training, as well as its ability to mitigate overestimation bias through the utilization of two critic networks. However, what sets our proposal apart is the integration of an AE within the main networks. This addition empowers our method to directly process high-dimensional observations, enabling the agent to learn directly from them. The complete architecture, including the number of layers and activation functions, is illustrated in Fig \ref{fig:Architecure}.

    \begin{figure}[ht]
        \centering
            \includegraphics[width=\linewidth]{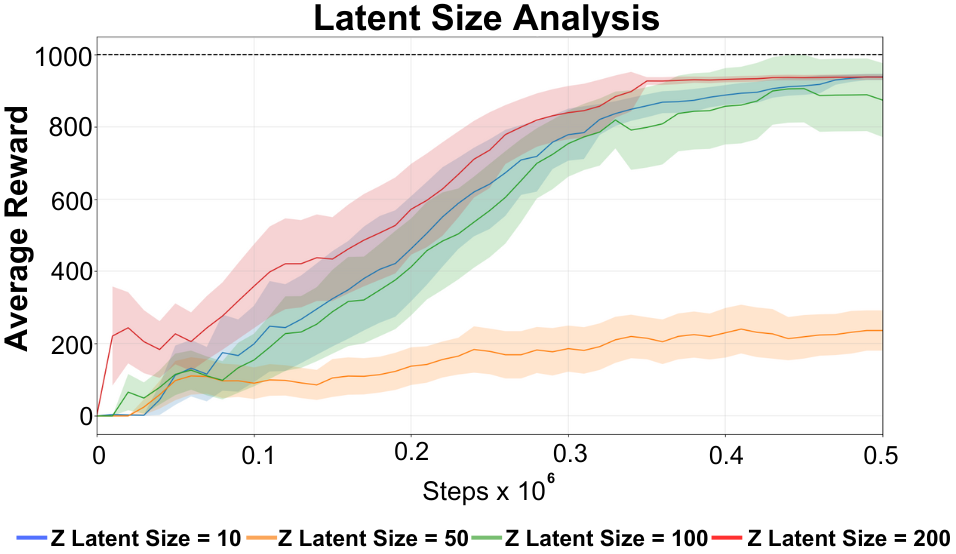}
            \caption{Optimal $z$ latent size analysis on the task of Ball and Cup. We ran a trial-error experiment under the same condition, changing the latent size to find the best value.}
            \label{fig:z_analysis}
    \end{figure}
    
    The Actor-Critic network learns and evaluates a policy. This network receives as input the low-dimensional vector $z$ from the encoder, which attempts to extract relevant features from the $s$ observations at time $t$. Here, both the encoder and decoder are trained by minimizing the $L2$ reconstruction loss between the input observation and the reconstructed observation. To determine the optimal latent space size, we conducted experiments training the agent under identical conditions while varying only the latent space size. The results of this analysis are presented in Fig. \ref{fig:z_analysis}. Based on these findings, a latent size of $z=200$ was selected for the remainder of the experiments. 
    
    In terms of policy learning methodology, we adhere closely to the established process of the base TD3 approach. However, instead of utilizing the true observation state $s$, we employ the latent representation vector $z$. Both the AE and the policy $\pi$ are trained simultaneously. Notably, the encoder and decoder networks employed for novelty detection are identical to those utilized for policy learning. A graphic representation illustrating how and where we update the gradients can be seen in the broken lines in Fig \ref{fig:Architecure}. The actor's gradients are not used to update the encoder networks, while the critic's gradients update the critic networks and the encoder at each training step. 
    
    One of the primary distinctions between our proposal and previous works, such as \cite{hafez2019deep} or \cite{yarats2021improving}, lies in the comprehensive update of the entire encoder at each step, encompassing both the convolutional and linear layers. We observed that this simple modification enhances the stability of the networks. We attribute this improvement to the fact that the linear layers play a crucial role in constructing the image representation. Furthermore, we maintain consistency by utilizing the same encoder for both the target and main networks (actor and critic) throughout the training process. Consequently, there is no requirement to update the encoder weights of the target networks at a faster rate than the rest of the network's parameters, as observed in \cite{yarats2021improving}.  

\subsection{Simulated Environments}

    \begin{figure}[ht]
        \centering
            \includegraphics[width=0.95\linewidth]{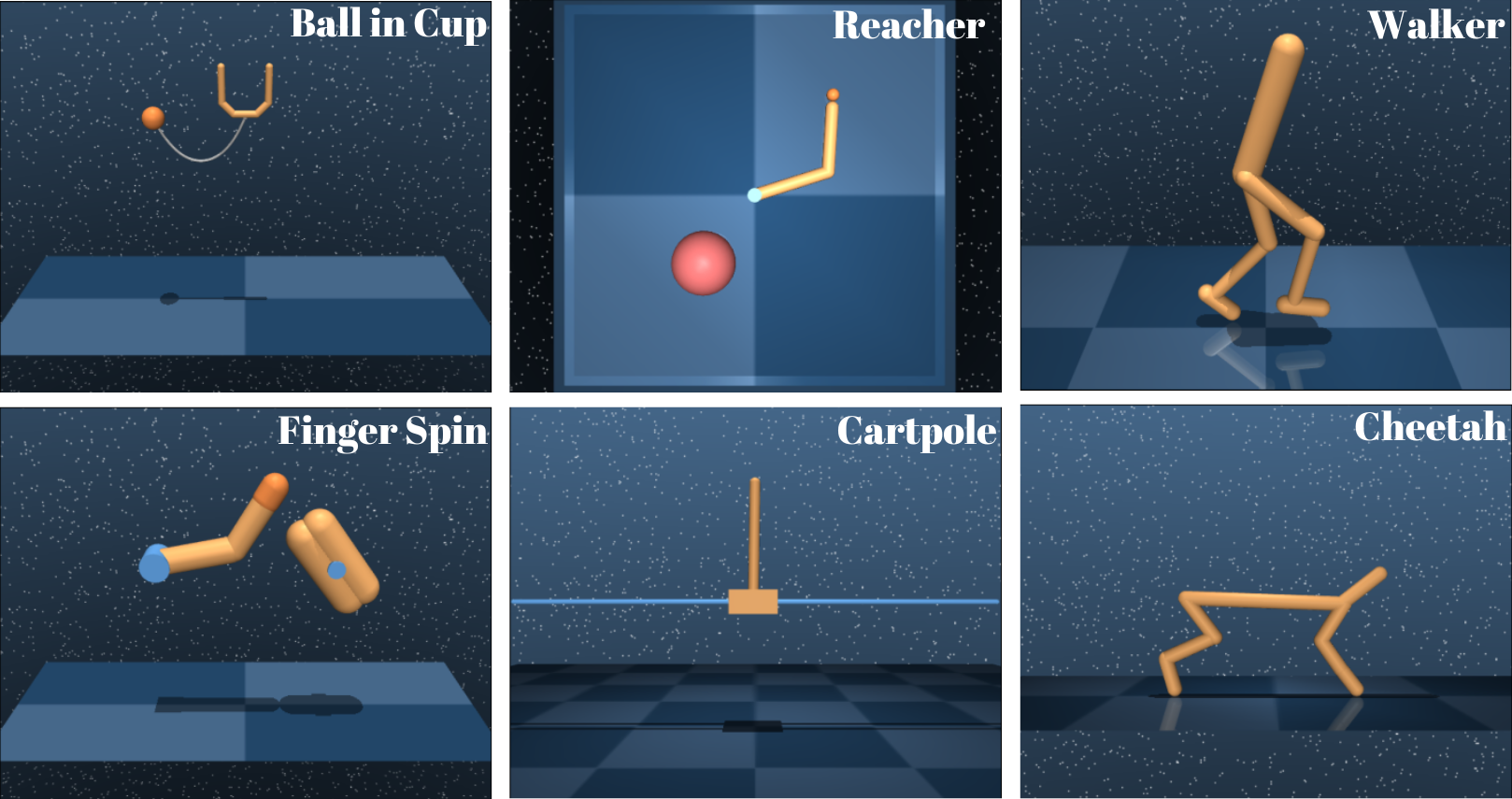}
            \caption{Image-based control tasks used in our experimentation. The \textit{\textbf{ball in the cup}} and \textit{\textbf{reacher}} tasks have a sparse reward that is only given once the ball is caught or the finger touches the red sphere, respectively. \textit{\textbf{Cartpole}} and \textit{\textbf{walker}} tasks require balance and constant movement. The \textit{\textbf{finger spin}} task includes contact between the finger and the object, while the \textit{\textbf{cheetah run}} task demands coordination and motion of a  significant number of joints. 
            }
            \label{fig:DMCS_Env}
    \end{figure}

    Repeating the nominal action is a popular technique used in RL. Several proposals \cite{lee2020stochastic, yarats2021improving, hafner2019learning, mnih2015human} claim that action repetition positively affects the learning process. However, suppose the number of repetitions is not set adequately. In that case, it can cause instability in the control dynamics and lead to sub-optimal behaviour \cite{yarats2021improving}, making choosing the correct number of repetitions a tuning hyperparameter that depends on each environment \cite{lee2020stochastic}. Likewise, repetition might not be suitable for real-world applications where servo-motors or linear actuators have fixed movements where action repetitions won't have any effect. Since our goal is an easily scalable algorithm capable of working in a wide range of applications, including real robots, we are not using this technique in our method; instead, we use novelty and surprise signals to obtain the same performance as the algorithms that use action repetition. More details of the training process as well as a list of hyperparameters, full source code and videos of each task can be found on the paper’s website at: \url{https://sites.google.com/aucklanduni.ac.nz/nasa-td3-pytorch/}
    
\section{Experimental Results} 
     We assess our algorithm's performance across diverse, complex continuous control robotics tasks in both virtual and real-world scenarios. Additionally, we conduct comparative analyses against SAC-AE \cite{yarats2021improving}, utilizing the official source-code implementation provided by the author \footnote{\url{https://github.com/denisyarats/pytorch_sac_ae}}. We choose challenging continuous control tasks of the DeepMind Control Suite \cite{tassa2018deepmind}. 
    
        \begin{figure}[ht]
        \centering
            \includegraphics[width=0.9\linewidth]{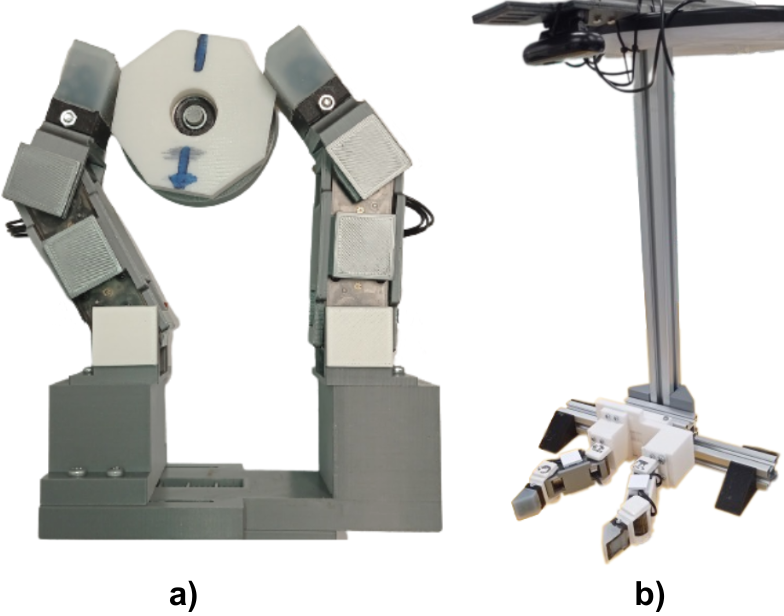}
            \caption{Subfigure a) presents the robotic gripper and the valve-like object that was 3D printed and attached to a fixed pivot or a servo-motor, depending on the test. Subfigure b) presents the low-cost robotic platform used.
            }
            \label{fig:Gripper}
    \end{figure}
    
    This test-bed platform provides complex environments for continuous control tasks with normalized actions and image state observations. We choose six tasks with different complexity and rewards; see Fig \ref{fig:DMCS_Env}. 
    A third-person camera image of size ($84\times84$) pixels with three channels is used for all tasks. A stack of three consecutive full-colour images is used as the state observation. We run each environment for $1\times10^6$ training steps with the same hyperparameters for five independently seeded runs. All parameters are trained with the Adam optimizer, and we perform $G=5$ (based on a trial error experimentation) gradient updates per environment step. We regularly evaluate the agent's performance in each environment during the training; after every $1 \times 10^4$ training steps, we compute the average reward over $10$ episodes. During the evaluation, no gradient updates are performed.

    
\subsection{Dexterous Robotics Manipulation in the Real World}
    To test the applicability and efficiency of our proposed method to solve real-world tasks, we describe an experiment involving a 4-DoF robot gripper \cite{valencia2023comparison}; see Fig \ref{fig:Gripper}. The gripper has two identical fingers equipped with Dynamixel XL-320 servomotors and a standard webcam (C270 Logitech with 720p) on top of the structure. The task requires turning a valve tap to an arbitrary target orientation. Despite the apparent simplicity, the dexterous manipulation task poses significant challenges due to its requirement for visual perception and physical coordination of 4-DoF. Specifically, the robot must learn to synchronize the movements of its two fingers to prevent cancellation of rotational movements, while also mastering a complex finger gait \cite{valencia2023comparison}. The resulting solution needs to account for real-world factors such as noise, control inaccuracies, and environmental conditions like lighting dynamics and friction. It is worth noting that we are conducting this task without the use of commercially expensive hardware platforms, and we aim to learn directly from raw images without modelling or system identification.
    
    The extrinsic reward for this task is computed as the normalized negative $L2-norm$ distance between the current valve's position and the desired angle. 
    The intrinsic reward is calculated with the novelty and surprise strategy mentioned above. The horizontal episode is limited to $50$ steps where we use an action space that outputs a four-element vector $\mathbb{R}^4$ with the desired position of each servomotor. We maintain the same image size as in the previous section to ensure consistency across experiments. Thus, the state space consists of the RGB images from the camera $(840\times640)$ downscaled to an image of size $(84\times84)$ pixels with three channels. 

    \begin{figure}[ht]
        \centering
            \includegraphics[width=\linewidth]{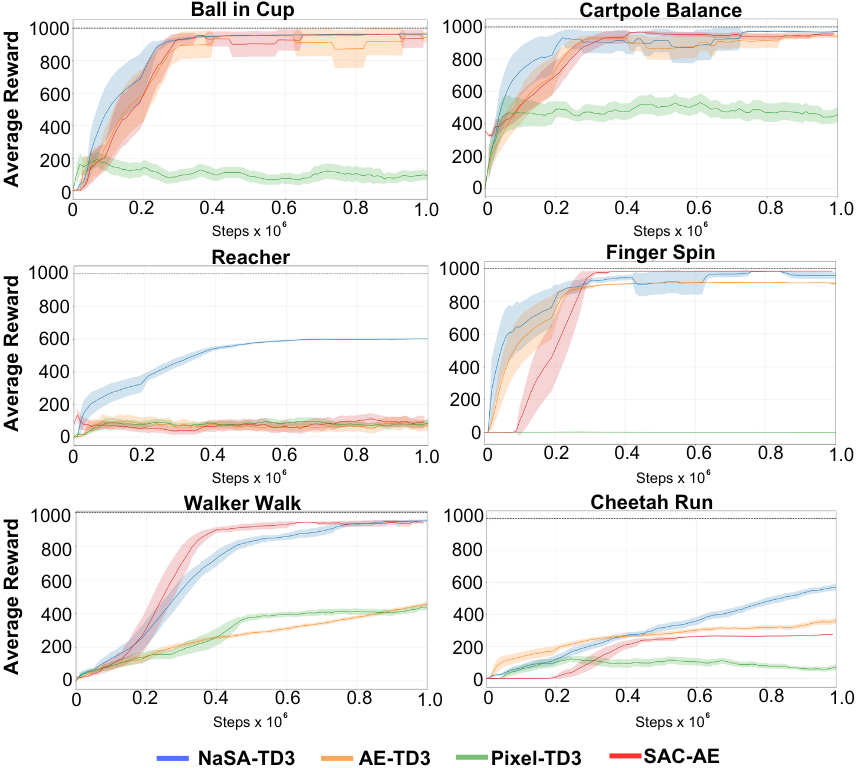}
            \caption{Learning curves for the 6 DeepMind Control Suite tasks.}
            \label{fig:simulation_Results}
    \end{figure}
    
    The robotic task has two scenarios at the end of the episode. In the first case, the valve orientation is left in place between episodes with no resetting or reinitialisation. In the second case, the valve orientation is reinitialised to a random angle (using a servomotor attached to the valve) once the episode ends. Previous related tasks rotate the valve to the same initial orientation at the end of each episode \cite{haarnoja2018soft,zhu2019dexterous,ahn2020robel,andrychowicz2020learning}. Reinitialising the valve to any random orientation between $[0, 2 \pi)$ increases the complexity of the task as it requires the agent to learn to use the information from the images to perceive the current valve's orientation. 
    We refrain from comparing this task with SAC-AE \cite{yarats2021improving} due to the necessity of modifying its original code. Our primary objective is to showcase that our proposal is readily applicable to real-world applications. We train the model for a total of $5 \times 10^4$ environment steps, which amounts to approximately 11 hours of real-world training time. Five independently seeded runs are used to confirm the stability of our proposal. All hyperparameters used in this experiment and task-solved videos are listed on our website.

\section{Results and Discussion }
    Fig. \ref{fig:simulation_Results} shows the average evaluation reward comparing our NaSA-TD3 proposal, TD3-AE (which is our image-based extension of TD3 without any intrinsic rewards), SAC-AE \cite{yarats2021improving} and Pixel-TD3 on the six continuous control tasks under identical conditions\footnote{For a fair comparison with our proposal with the exact number of steps for an episode, we present the result of SAE-AE passing action repetition parameter equal to 1}. Utilizing the novelty and surprise signals as intrinsic rewards significantly enhances task performance, with instances of achieving maximum reward values (corresponding to a $1 \times 10^3$ max return) observed in certain environments. In contrast, some other image-based approaches fail to reach the maximum value or struggle to solve the task entirely, as exemplified by Pixel-TD3. 

        \begin{figure}[ht]
        \centering
            \includegraphics[width=\linewidth]{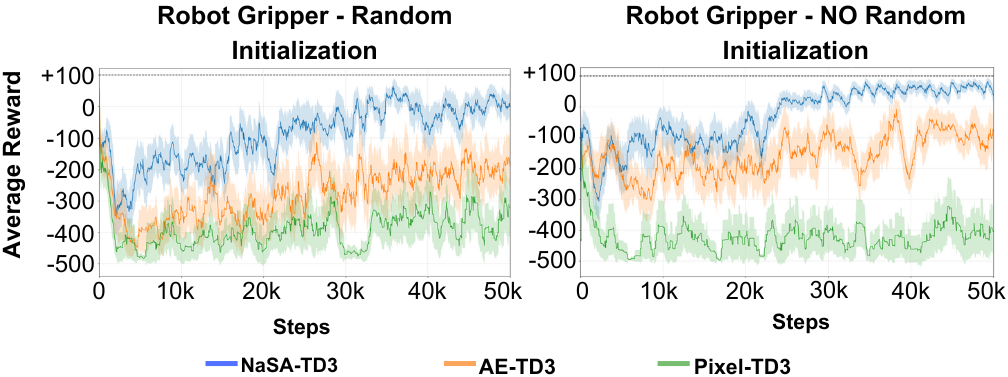}
            \caption{Performance curves of NaSA-TD3, AE-TD3 and Pixel-TD3 on the real robot gripper with a valve rotation task. Training end-to-end on the real robot for both scenarios.}
            \label{fig:Gripper_results}
    \end{figure}
    
    It is worth noting that AE-TD3’s performance can match or improve SAC-AE in some environments. By further incorporating novelty and surprise stimuli as intrinsic rewards, NaSA-TD3 outperforms all variations. Notably, our method does not require a complex tuning of hyperparameters or complicated training steps as other image-based RL algorithms. Nevertheless, in the spirit of transparency regarding the results, it is worth noting that the novelty and surprise signals do not significantly impact the final results when the task is not highly complex, or the extrinsic reward is not sparse. This can be seen more clearly in environments such as \textit{ Walker} or \textit{Finger Spin}, where the agent performed similarly to SAC-AE. Presumably, this is because the intrinsic reward is not required, implying that the problem is easy enough to be solved with an extrinsic reward. In contrast, when the environment is complex or has sparse rewards, as in the \textit{Reacher} or \textit{Cheetah} environments, the intrinsic signals help significantly improve the performance.
    
    Fig \ref{fig:Gripper_results} shows the average evaluation reward for the real-world manipulation experiments. The results show the effect of novelty and surprise signals in executing the dexterous manipulation task. The superiority of our proposal over extrinsic-reward-only methods is even more evident. Our proposal solves the task and gets the highest reward in both tested scenarios. Pixel-TD3 failed to solve the tasks, while AE-TD3 struggled when the valve was reinitialised to a random orientation. These results underscore the effectiveness of intrinsic stimuli in encouraging the agent to explore a broader range of state-action pairs in complex environments, ultimately leading to faster and improved performance in real-world scenarios.
    
    Our proposal does have some limitations, with one of the most significant being the amount of RAM required. Utilizing pure images as input and storing them in a replay buffer can be computationally demanding, necessitating at least 64 GB of RAM to run a simulation environment for 1 million environment steps. However, this challenge can be mitigated by tuning the buffer size. Our experimentation revealed that other related proposals, such as \cite{yarats2021improving} and \cite{lee2020stochastic}, require similar or even greater amounts of RAM, as they often train models for more extensive training steps. Notably, this issue is not explicitly addressed in their papers. Furthermore, despite our method demonstrating superior and faster convergence, the computational time for each environment step is higher compared to other methods such as SAC-AE \cite{yarats2021improving}. This is primarily due to the additional gradient updates for the autoencoder and policy in each step, as well as the computation of  SSIM and MSE for novelty and surprise values, which involve additional calculations and resources.
    
\section{Conclusion}
    In this work, we set out to investigate the use of intrinsic rewards as a mechanism for improving the performance of RL in the execution of complex tasks such as dexterous robotic manipulation. We formulated reward metrics inspired by novelty and surprise, using the predictability of patterns across image sequences. A key novelty was the use of Autoencoders, which reduced our system's computational and memory needs to support real-world computation. Our findings show the ability to learn directly from raw images in simulation and real-world robots. This opens the door to scaling this proposal to more complex scenarios and tasks where the only information needed can come from a camera. Likewise, we verify that when the reward is sparse, auxiliary intrinsic signals such as novelty and surprise can help to discover efficient exploration strategies and assist the learning process.
    
    Future work will seek to include other intrinsic signals such as boredom, frustration, and pleasure. Furthermore, we will seek to improve the efficiency in the number of samples necessary, perhaps with the adaption of model-based reinforcement learning algorithms, plus a more comprehensive analysis of other autoencoder architectures and their effect on policy learning.

\section*{Acknowledgements}
    This research was supported by New Zealand's Science For Technical Innovation (SfTI) on contract UoA3727019. We want to thank the Human Aspects in Software Engineering Lab (HASEL) from the University of Auckland. 

\bibliography{bibliography}
\bibliographystyle{ieeetr}

 
\end{document}